\documentclass[11pt]{article}
\usepackage{coling2018}
\usepackage{times}
\usepackage{url}
\usepackage{latexsym}
\usepackage{comment}
\usepackage{graphicx}

\title{AI4D - African Language Dataset Challenge}

\author{Kathleen Siminyu, Sackey Freshia, Jade Abbott, Vukosi Marivate \\
  Masakhane, Africa \\
  masakhane.io \\
  {\tt masakhane-mt@googlegroups.com} \\} 
  
\author{Kathleen Siminyu \\
  Artificial Intelligence for Development \\
  Africa \\
  {\tt kathleensiminyu@gmail.com} \\\And
  Sackey Freshia \\
  Jomo Kenyatta University \\
  of Agriculture and Technology \\
  {\tt freshiasackey@gmail.com} \\\AND
  Jade Abbott \\
  Retro Rabbit \\
  {\tt jabbott@retrorabbit.co.za} \\\And
  Vukosi Marivate \\
  University of Pretoria \\
  {\tt vukosi.marivate@cs.up.ac.za} \\}
  
\begin{document}
\maketitle
\begin{abstract}
As language and speech technologies become more advanced, the lack of fundamental digital resources for African languages, such as data, spell checkers and Part of Speech taggers, means that the digital divide between these languages and others keeps growing. This work details the organisation of the AI4D - African Language Dataset Challenge\footnote{https://zindi.africa/competitions/ai4d-african-language-dataset-challenge}, an effort to incentivize the creation, organization and discovery of African language datasets through a competitive challenge. We particularly encouraged the submission of annotated datasets which can be used for training task-specific supervised machine learning models.
\end{abstract}

\section{Introduction}

Africa has a language diversity of over 2000 languages, many of which are only spoken and not written \cite{enthnologue}.

As language technologies advance and more sophisticated tools are built using Artificial Intelligence, the divide between low resource languages and others is likely to get even larger as a common prerequisite of these advanced systems is the existence of a large amount of digital data. African languages, low resource languages, are at risk of being left behind \cite{Joshi2020TheSA}.

Data Science and Machine Learning skills are increasingly becoming widespread on the African continent. This can be attributed to the rise of grassroots capacity building efforts through organisations such as Data Science Africa\footnote{http://www.datascienceafrica.org/}, Data Science Nigeria\footnote{http://datasciencenigeria.ai}, Deep Learning Indaba\footnote{https://deeplearningindaba.com/}, as well as NLP-specific communities such as Masakhane\footnote{https://www.masakhane.io/} \cite{orife2020masakhane}. These movements have facilitated a critical mass of individuals with the relevant skills, who speak African languages that can start contributing to the overall body of work that currently exists and begin the work where none does. With this challenge, we sought to engage the African NLP community in the task of dataset creation.

\section{Methodology}

The realisation of this work was wholly driven by the intended outcome, the need for more African language datasets for use in NLP research. 

\subsection{Framing the Challenge}

Early framing of this challenge was predicated on the fact that pre-trained language models are producing state-of-the-art NLP results \cite{devlin2018bert,radford2019language}. African NLP would undoubtably benefit from the creation of such language models, and so the competition was initially envisioned as a language model challenge. Unfortunately, not only do most pre-trained language models require large amounts of monolingual data to train, they require labelled NLP datasets in order to usefully evaluate the models. This motivated the decision to create a challenge focused on data collection, rather than model building.

\subsection{Securing Donor Funding}
\label{sect:pdf}

With the aim of creating datasets that are openly available, we prepared and circulated a proposal and succeeded in securing donor funding. That being committed, we tailored the challenge to take place in 2 phases. The initial phase focused on data collection and the second phase being a more conventional machine learning (ML) challenge, where the datasets developed in the first phase could be used as evaluation sets for a pre-trained language model challenge.

\subsection{Hosting the Challenge}
\label{ssec:layout}

Our target audience was researchers, practitioners and enthusiasts from African countries who could create datasets for the languages that they speak. We approached Zindi\footnote{http://zindi.africa}, an African data science competition platform, to host the challenge in a bid to leverage their existing user base. They have over 12,000 individuals signed up.

\subsection{Evaluation of Submissions}
Unlike a conventional ML challenge that would have an agreed upon automated metric to evaluate and rank submitted models, evaluation of a challenge of this kind cannot be automated. Instead, we put together a panel of judges with experience in NLP who would review the datasets each month. We also indicated in the challenge guidelines that each dataset submission should be accompanied by a datasheet \cite{gebru2018datasheets} that documents its motivation, composition, collection process, recommended uses, and so on (Example in Appendix A). 

Evaluation was done by judges analysing the datasheets and awarding points to each submission based on a scoring rubric. The rubric took into account the following: how representative and balanced the corpus was, the dataset size in terms of tokens and unique tokens, whether it was annotated for a specific downstream task, under-representation of the language in terms of digital data, methodology of the data collection and labelling process, originality of the data collection and labelling process (Reviewer Documentation in Appendix B).

\section{Results}
\label{sec:length}

\begin{table}[h]
\begin{center}
\begin{tabular}{|l|r|l|}
\hline \bf Language & \bf Tasks & \bf Submissions \\ \hline
Yoruba & MT, Diacritic Verification, Text Classification, NER, misc & 7  \\
Kiswahili & Document Classification, misc & 6  \\
Igbo & NER, misc & 4 \\
Hausa & Sentiment Analysis, Document classification, misc & 4  \\
Fongbe & MT, Speech to Text, misc & 3  \\
Amharic & Hate speech detection, stop words list, misc &  3 \\
Asante Twi & Sentiment Analysis, MT, misc & 3 \\
Chichewa & NER, MT & 2 \\
Ewe & MT, misc & 2 \\
Wolof & ASR & 1\\
Tunizian Arabizi & Sentiment Analysis & 1 \\
Kikuyu & misc & 1 \\
Kabiwe & MT & 1 \\
Oromo & misc & 1 \\
Zulu & misc & 1 \\
\hline
\end{tabular}
\end{center}
\caption{\label{font-table} Language and Task distribution of submissions. }
\end{table}

The challenge ran for a period of five months with a total of 270 people registering on the Zindi platform to participate.

The data in the submitted datasets was compiled from a wide variety of sources. These were largely digital sources such as news websites, religious texts, Facebook, Twitter and YouTube.  This outcome is likely an indication of the ease of access that online sources present, given that the data is already digitised. Other data creation processes included participants convening to carry out manual translation of existing pieces of text.  Unique contributions included OCR to digitize printed texts and using the user base of a commercial application to crowd source and validate recordings of phrases and texts common on the platform. 


{\bf Observations and Lessons Learned}:
\begin{itemize}
\item Teams composed of individuals from relevant multi-disciplinary backgrounds, including computer scientists, professional translators and linguists, were able to create and annotate datasets that captured fundamental lexical and semantic nuances of languages. 
\item The challenge framing allowed for anyone to participate. While useful as an exercise in evaluating the interest in such a challenge, the top evaluated submissions came from teams who had been exposed to NLP research work. Targeting such a challenge to NLP researchers could lead to higher quality submissions in future. 
\item A portion of submissions contained very few data points. As the aim is to use the datasets for NLP, in future, we'd set explicit minimum requirements  with regards to the size of datasets admissible, file formats and require the inclusion of any cleaning or pre-processing code used.
\item Since the challenge was evaluated monthly, we often received disparate submissions from the same teams as they managed to obtain more data. Instead, one large dataset built over a couple of months would have been the ideal outcome, so in future we'd select and support teams for a sustained period of time to enable them build sizeable datasets.
\item  Participants and judges had to rely on their own understanding of what "Representative and Balanced" means in a dataset. In future, more specificity of what "representative" and "balanced" means would enable participants to produce better datasets.
\end{itemize}

\section{Future Work}
A large number of opportunities were identified to support future work in African, and low resource, language dataset creation, as follows:

\begin{itemize}
\item Research and analysis of the legal implications of obtaining textual, visual and audio data from a variety of online sources, which were noted as a common source among participants. The copyright and intellectual property implications will have to be thoroughly assessed ahead of the publication and further public use of relevant datasets.
\item Outlining of best case practise techniques for protecting the identities and privacy of users, in instances where data is obtained from social media/content platforms like
Twitter, Facebook and YouTube. Social media sites have been noted as a common data
source.
\item Recommended techniques for identifying and ascertaining whether data obtained from
online sources(news publications,social media and content platforms) contains biased
sentiments(sexist, racist) and offensive material(hateful), as well as techniques for removing any biased sentiment and offensive material, if need be, depending on the use of the dataset.
\end{itemize}

Courtesy of this dataset creation challenge, we have secured further funding to support 5 of the top teams for a 6 month period. During this time, they will further flesh out their datasets with the aim of using these as the basis of future NLP challenges/shared tasks. This project will also be used as a model case to inform evidence-based policy making concerning Artificial Intelligence and we hope that it will be replicated to support the development of data for other low resource languages.

\section*{Acknowledgements}

This work has been funded through a partnership between the International Development Research Centre, the Swedish International Development Cooperation Agency, the Knowledge4All Foundation, Zindi Africa and the AI4D-Africa Network. The expert panel that volunteered their time to undertake the difficult qualitative task of dataset assessment was composed of Jade Abbott - Retro Rabbit, John Quinn - Google AI / Makerere University, Kathleen Siminyu - AI4D-Africa, Veselin Stoyanov - Facebook AI and Vukosi Marivate - University of Pretoria. 

\blfootnote{
    %
    %
    %
    %
     \hspace{-0.65cm}  
     This work is licensed under a Creative Commons 
     Attribution 4.0 International Licence.
     Licence details:
     \url{http://creativecommons.org/licenses/by/4.0/}.
    %
    %
}

\bibliographystyle{acl}
\bibliography{acl}
\raggedbottom

\section*{Appendix A: Copy of Outstanding Submission Datasheet}

\includegraphics[scale=1, trim=2cm 0 0 3cm]{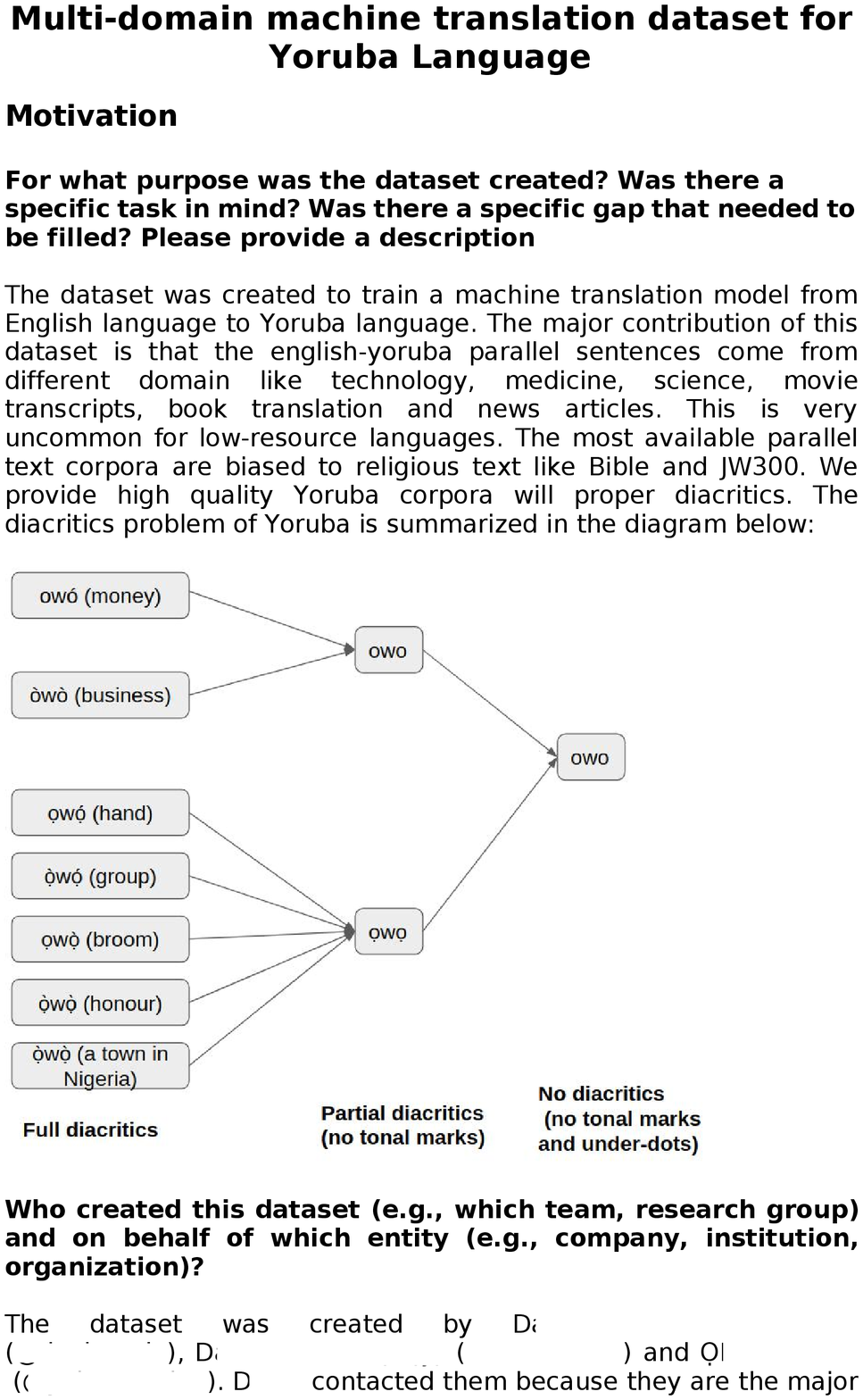}
\includegraphics[scale=1, trim=2cm 0 0 3cm]{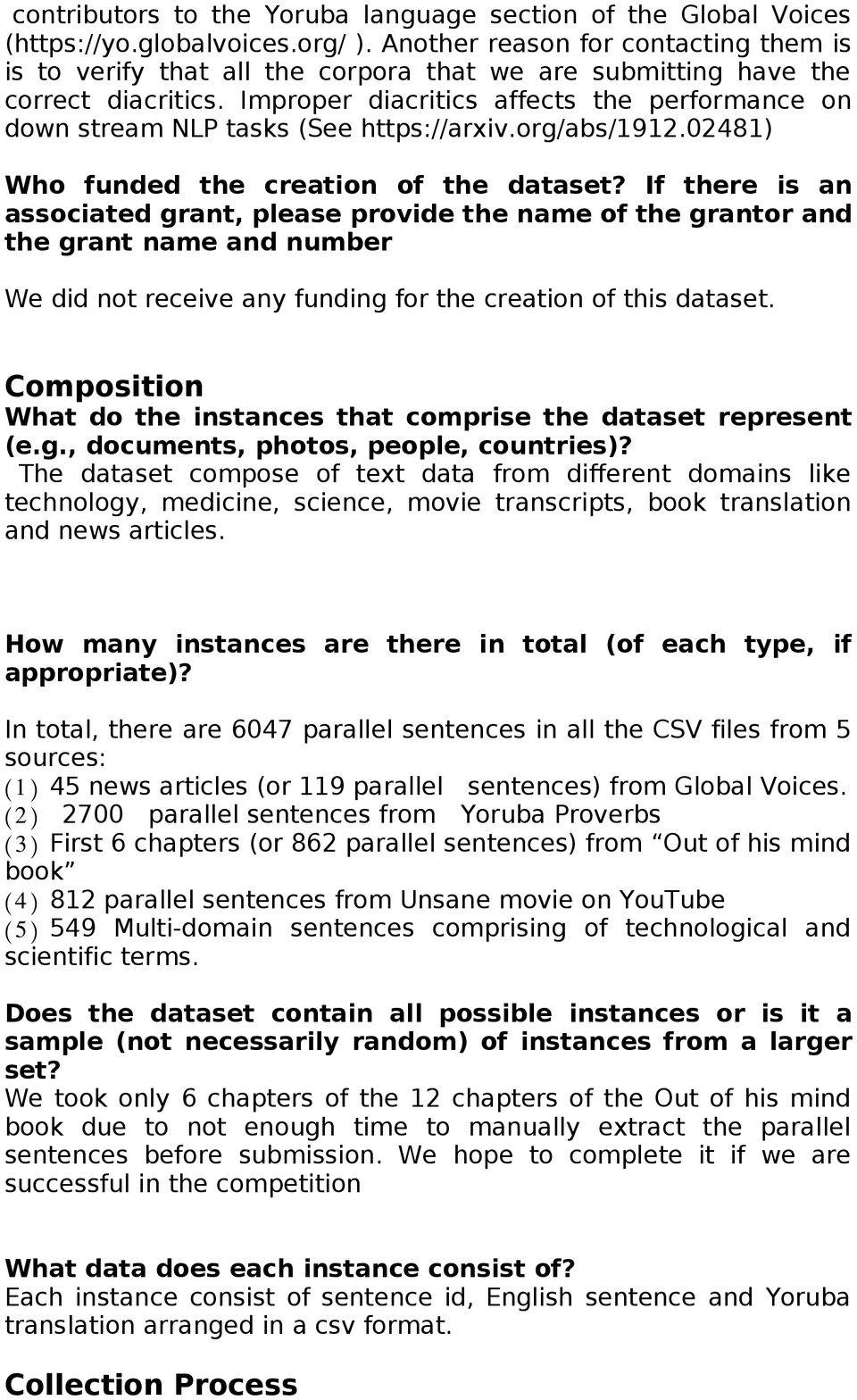}
\includegraphics[scale=1, trim=2cm 0 0 3cm]{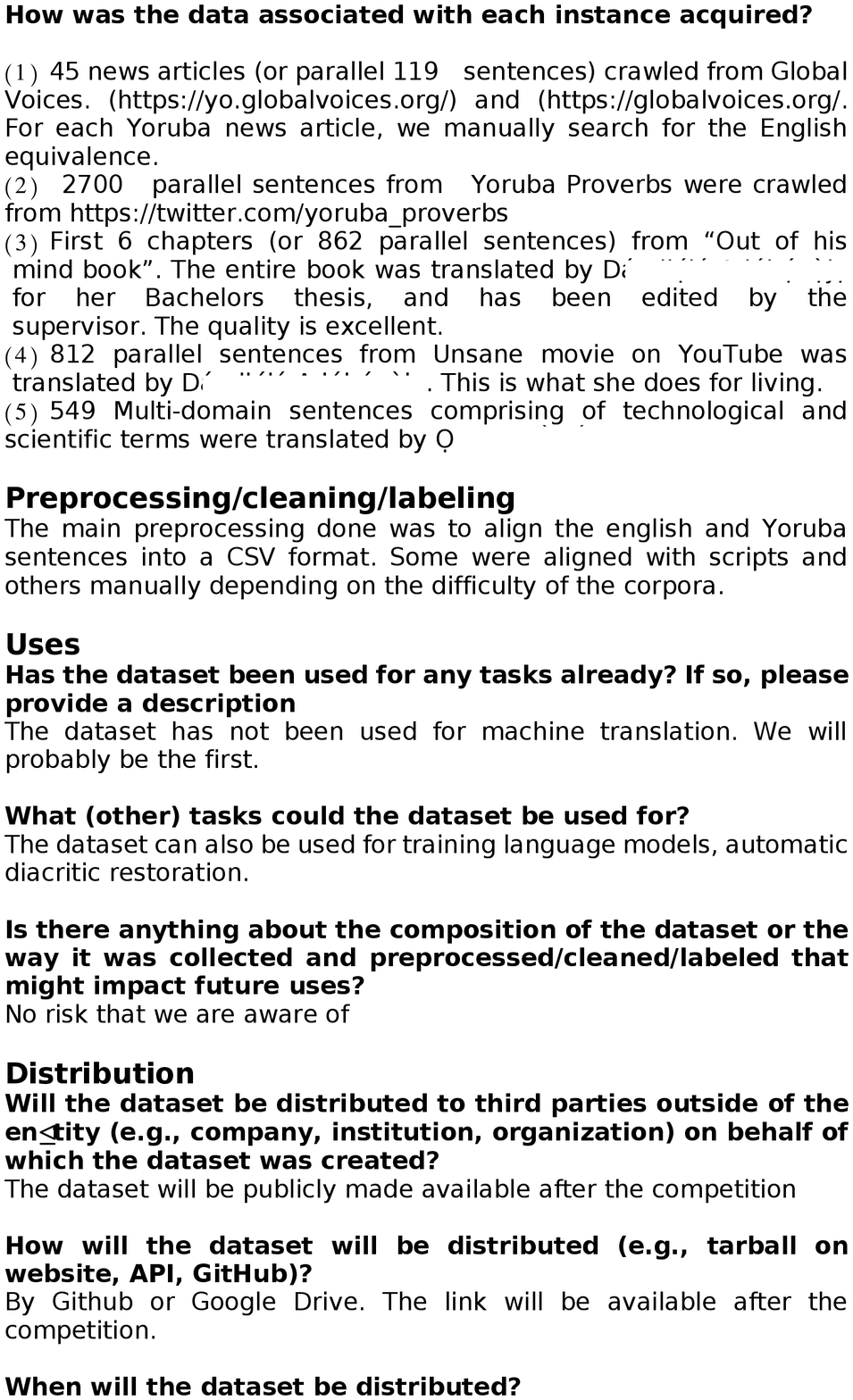}
\includegraphics[scale=1, trim=2cm 0 0 3cm]{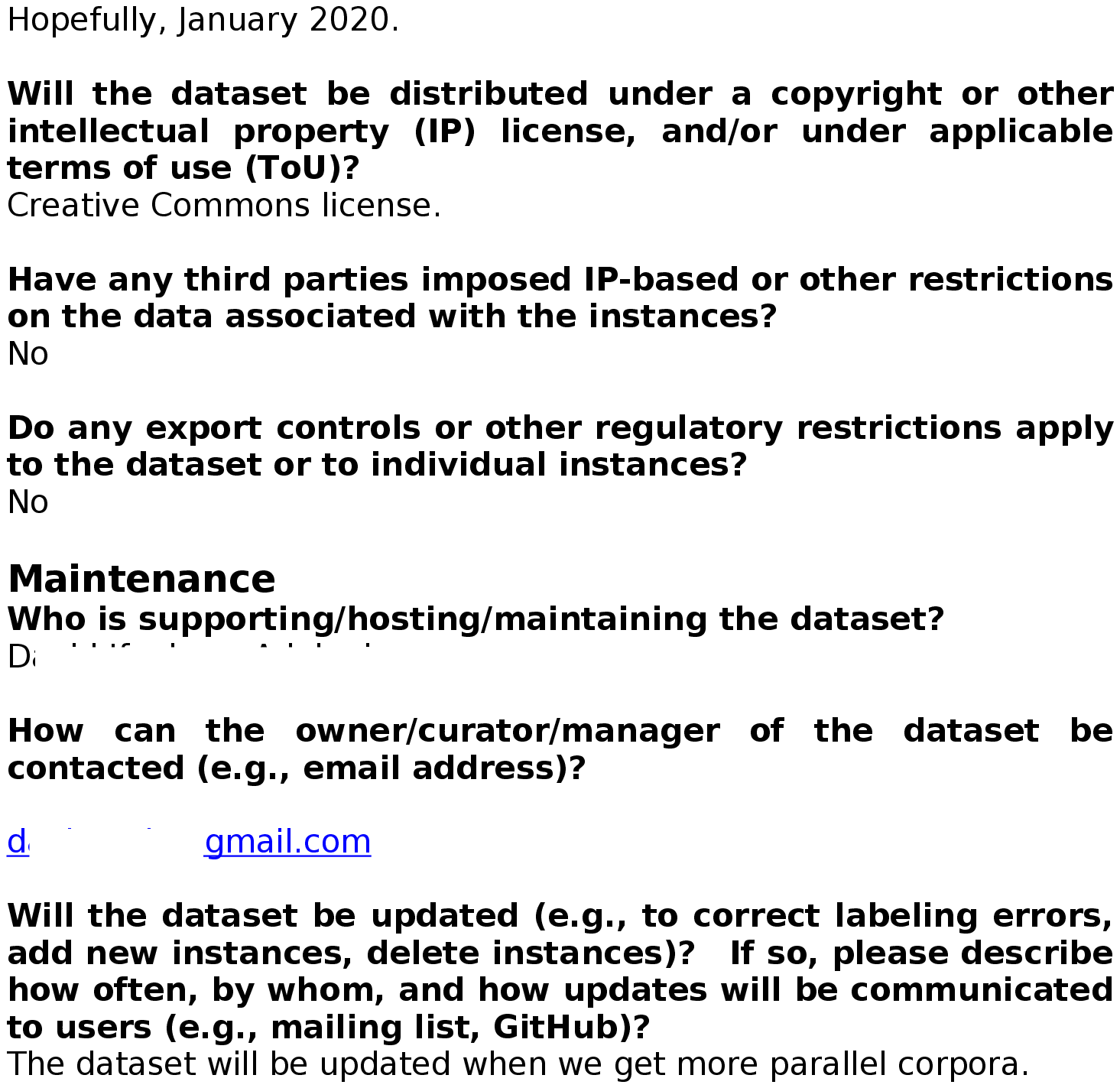}

\section*{Appendix B: Copy of Reviewer Documentation}
\includegraphics[scale=1, trim=2cm 0 0 3cm]{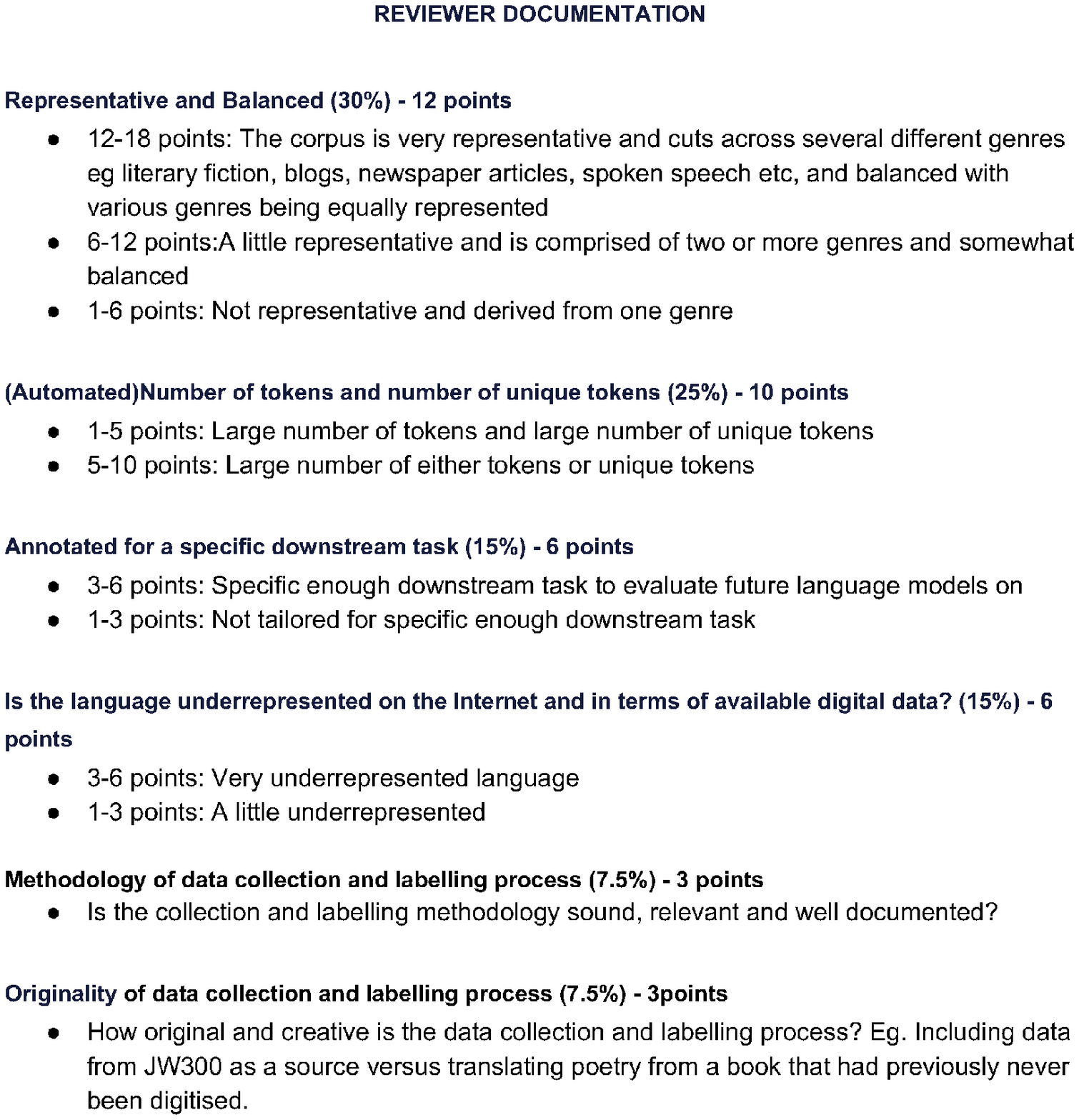}

\end{document}